\documentclass[runningheads]{llncs}

\usepackage{accvabbrv}

\usepackage{graphicx}
\usepackage{booktabs}

\usepackage[accsupp]{axessibility}

\usepackage{graphics}
\usepackage{xcolor}
\usepackage{subcaption}
\usepackage[latin1]{inputenc}
\usepackage{amssymb,amsmath,array}
\usepackage{gensymb}

\usepackage{multirow}
\usepackage{colortbl}
\usepackage{array}
\usepackage{ragged2e}

\usepackage{svg}

\usepackage{hyperref}

\usepackage{orcidlink}

\begin{document}

\title{Few-shot target-driven instance detection based on open-vocabulary object detection models} 

\author{Ben Crulis\inst{1}\orcidlink{0000-0002-3132-1136} \and
Barthelemy Serres\inst{1}\orcidlink{ 0000-0003-3702-1334} \and
Cyril De Runz\inst{1}\orcidlink{ 0000-0002-5951-6859} \and
Gilles Venturini\inst{1}\orcidlink{ 0000-0002-8112-2418}}

\authorrunning{B.~Crulis et al.}

\institute{University of Tours, Blois, France 
\email{\{ben.crulis,barthelemy.serres,cyril.derunz,gilles.venturini\}@univ-tours.fr}}

\maketitle

\begin{abstract}
Current large open vision models could be useful for one and few-shot object recognition. Nevertheless, gradient-based re-training solutions are costly. On the other hand, open-vocabulary object detection models bring closer visual and textual concepts in the same latent space, allowing zero-shot detection via prompting at small computational cost.
We propose a lightweight method to turn the latter into a one-shot or few-shot object recognition models without requiring textual descriptions. Our experiments on the TEgO dataset using the YOLO-World model as a base show that performance increases with the model size, the number of examples and the use of image augmentation.
\end{abstract}

\section{Introduction}
\label{sec:intro}
In the recent years, the field of computer vision has quickly progressed and now proposes accurate, lightweight and efficient object detectors such as YOLOv7~\cite{wang2023yolov7}. However, these detectors are trained for fixed sets of classes they can detect. While these object detectors can be re-trained to detect new classes using Transfer Learning~\cite{zhu2023transfer} or Meta-Learning~\cite{NajdenkoskaZW23}, this method requires large datasets and a costly fine-tuning step. To alleviate these limitations, recent research efforts now permit the creation of models that allow users to choose which new classes the model should detect after training has stopped, by specifying these classes using natural language. Models with this property are called Open-Vocabulary models~\cite{minderer2022simple,Cheng2024YOLOWorld}.

However, specifying new classes using natural language is still limited in some contexts, for instance when the objects are not easily described using language. In addition to this, there is also a need for personalized object detectors that can recognize new objects instances that belong to or are of some particular interest to the user, especially for visually impaired people, our experimental context.

To address this issue, the paradigm of Target Driven Instance Detection (TDID) was created, but has not gained much attention over the recent years. In the TDID paradigm, one or more target images are submitted to the model to specify the objects that have to be detected in a query image. The attractiveness of TDID is that models adopting this paradigm should require only one or a few example target images to detect the objects. This property makes this approach a particular case of Few-Shot Learning~\cite{wang2023recent}.

In this work, we propose to avoid training a TDID model from scratch or fine-tuning a pre-trained model and instead adapt an existing open-vocabulary detection model to turn it into a TDID model using very little computation. This approach is motivated by the high energy and economical costs of training deep learning models using stochastic gradient descent optimization. This presuppose the original model is pre-trained and sufficiently powerful to handle such a radical conversion process. In the next sections we describe the approach and then evaluate its performance. To our knowledge, this is the first attempt of this kind in the literature.


\section{Related Work}

The Target Driven Instance Detection paradigm \cite{ammiratoTDID18,Mercier2021} consists in giving the model one or more example images, called target images, of the new object or class the model should detect in a query image where the object might be present or not.

\subsection{YOLO-World}

YOLO-World \cite{Cheng2024YOLOWorld} is a real-time, open vocabulary model trained to detect objects given a prompt that is a list of classes specified in natural language and inserted in the Re-parameterizable Vision-Language
Path Aggregation Network (RepVL-PAN) part of the model. This gives zero-shot detection abilities to the model. The \textit{prompt-then-detect} approach of YOLO-World allows it to detect several objects in a single forward pass, making it very efficient at inference time. The reparametrization ability of the RepVL-PAN module is crucial to enable the conversion of the whole model into a TDID model.

The YOLO-World architecture is made of a convolutional neural network model backbone which extract image embeddings for the detection, and of an object detector head containing the RepVL-PAN module which takes image and text embeddings to make classification and bounding box predictions. The RepVL-PAN module accepts CLIP \cite{Radford2021LearningTV} embedding vectors as inputs. These vectors are normally obtained by using the text encoder part of CLIP. However, the CLIP model is constituted of two modules, the text encoder and the image encoder that respectively map text and images to a common latent space with $512$ dimensions. The two modules of the pre-trained CLIP model were trained in an end-to-end manner with a contrastive loss to make matching image-text pairs have a high dot product in the latent space and a small dot product for mismatched pairs. Images and text representing similar visual and semantic concepts should thus be encoded to similar vectors in this space.

\subsection{Other Object Detection Models Supporting Prompting}
Other recent models have similar architecture as YOLO-World, for instance, the Segment Anything Model (SAM) \cite{kirillov2023segment} is a vision model with several modes of prompting including an open vocabulary prompting mode. Conveniently, SAM also leverages the CLIP text encoder for the natural language prompting mode, which also makes it a candidate for our TDID conversion method. However, unlike YOLO-World, SAM is specialized in image segmentation and is not aimed at a real-time usage.

The T-Rex2 model \cite{jiang2024trex2} on the other hand natively supports visual prompting to detect objects similar to a target object. While this method already falls into the TDID paradigm without conversion, this model requires an additional user-specified bounding box and does not allow several target objects to be specified at the same time in the prompt, which makes it slightly different from our method. Finally, the weights of the T-Rex2 model are private and inference is only possible through an API.

The Deep Template Instance Detection method \cite{Mercier2021} is another work that proposes to re-detect objects based on instance templates, so their approach belong to the TDID paradigm. They specifically train a deep neural network to match object templates acquired from different viewpoints to ground truth detections of the same object instance using a synthetic dataset. Their architecture uses both local and global templates to separate general and viewpoint specific features, to simplify the work of the model. The requires template images to be segmented from their background, which is made possible thanks to the synthetic nature of their training dataset.

More recently, Large Language Models (LLMs)~\cite{raiaan2024review} also have been augmented with vision capabilities~\cite{zhang2024vision,liu2023llava}. In particular, Florence-2~\cite{xiao2023florence} was trained to have the capability to do open-vocabulary object detection, along with other vision capabilities. For this model, the input image is tokenized into image embeddings and become part of the prompt of the LLM along with the token specifying the task and other possible complementary information. The model then decodes the output tokens to produce text and position information of bounding boxes. However, as all LLMs, this models requires several forward passes to auto-regressively predict all tokens, making this approach a priori less efficient for object detection.
In the case of open-vocabulary detection, the input classes are specified in plain text and tokenized, instead of accepting embeddings from a shared text-image latent space, so this model cannot be converted with the method described in the next section.

\section{Transforming YOLO-World into a Target Driven Instance Detector} \label{method}

\begin{figure}[t]
\centering
\includegraphics[scale=1.0]{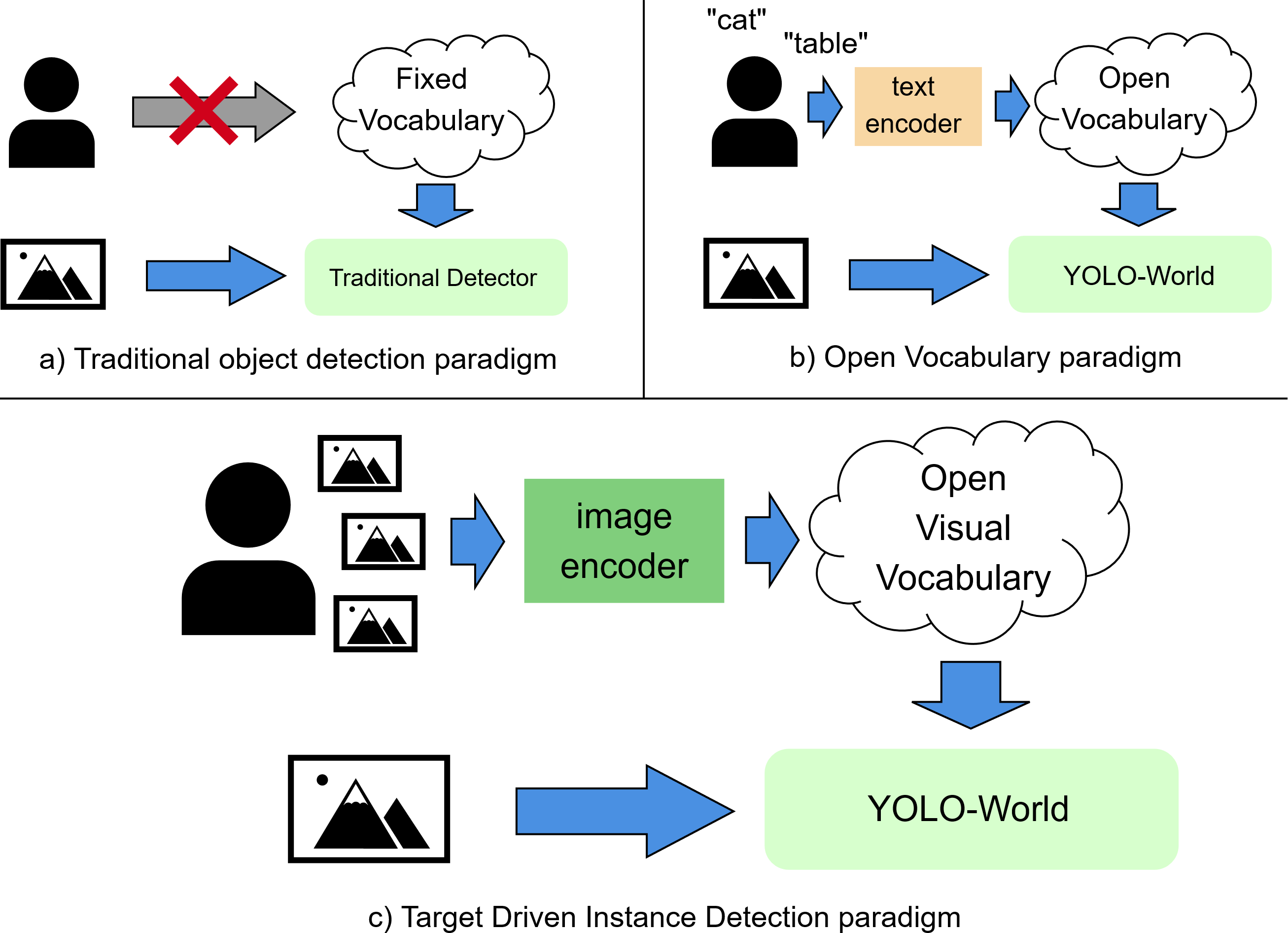}
\caption{Overview of the different detection paradigms. Our proposed method falls into the TDID paradigm (c). The traditional object detection paradigm (a) produces models with a fixed set of object classes that can be recognized, which gives no control over the objects that can be detected or not to the user. The Open Vocabulary paradigm (b) gives some control to the user by allowing them to use natural language text prompts that specify the classes the model should detect. Finally, the Target Driven Instance Detection paradigm (c) proposes to directly use one or more image examples of the target object as prompt for the model so that it can detect the object in other images.}\label{Fig:overview}
\end{figure}

\begin{figure}[t]
\centering
\includegraphics[scale=0.85]{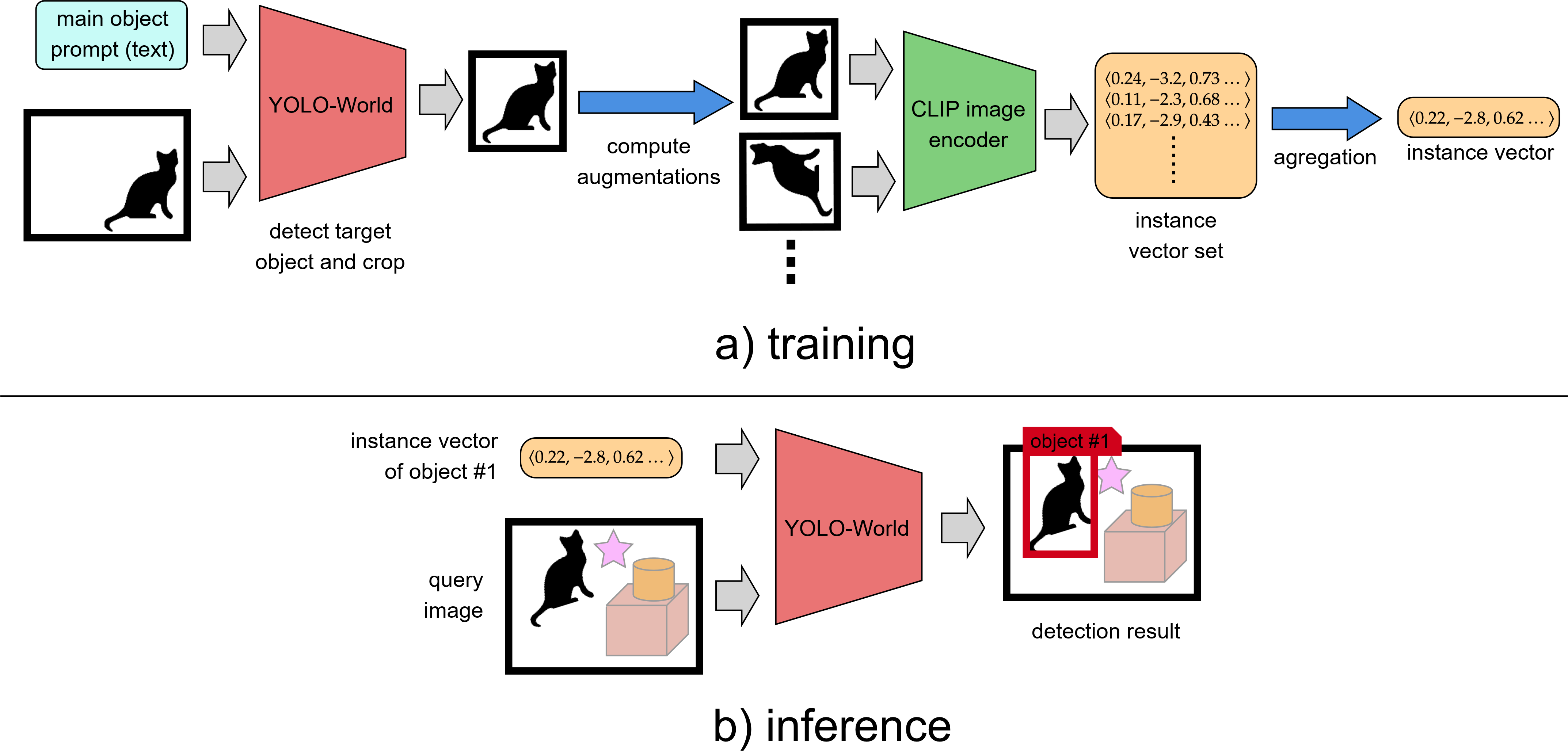}
\caption{
Overview of the proposed method. In the training phase (a), YOLO-World is used with a text prompt such as \textit{"main object"} to detect the most salient object in the image and use the the inferred bounding box to crop the image to the object. The cropped image is then optionally augmented and the resulting embeddings are aggregated to get a single vector per saved object. In the inference phase (b), the saved object embeddings are used as prompt in YOLO-World, which allows detecting the saved objects in other images. 
}\label{Fig:ourmodel}
\end{figure}

This section demonstrates how to turn an Open Vocabulary object detector that accepts semantic embeddings into a TDID. We propose a method that does not require any fine-tuning and which allows to recognize new object instance using as few as one example image per target object. At its core, our method consists in changing the provenance of the input embeddings from text embeddings to image embeddings, making this approach similar to \textit{prompt-learning}~\cite{miyai2024locoop}. A summary comparison between the different paradigms can be seen in Figure \ref{Fig:overview}.

The YOLO-World model accepts two different inputs, an image $x \in \mathbb{R}^{H \times W \times 3}$ where $H$ and $W$ are the height and the width of the image respectively and a list of $c$ semantic embeddings $z_i \in \mathbb{R}^d$ where $d=512$ and $i \in [1,c]$, that represents the $c$ classes the model should detect in $x$. $d$ is set to $512$ because of the pre-trained CLIP model which encodes text and images in a shared $512$-dimensional latent space.

\subsection{Model Conversion}

The main idea of this work is that, instead of computing $z_i$ from text in natural language to specify the classes using the text encoder part of CLIP as in the YOLO-World paper, we compute $z_i$ from one or more images using the image encoder part of CLIP. More specifically, we want the user to provide images of the target object clearly visible and convert these images into a vector latent space of CLIP so that the vector characterizes this object. 

As the target object may not occupy the entire image, we first apply a detection step to crop the object and separate it from the background and other irrelevant objects. This is achieved by calling YOLO-World on the target image with a special prompt that will tell the model to detect the most prominent object in the image. In this work, we empirically find that the following prompt works: ``main object''. We run the Non-Maximum Suppression algorithm (NMS) on the YOLO-World output using this prompt to find the single most likely detection bounding box. The inferred bounding box is then used to crop the original image with some user-specified margin.

Then, in order to improve the quality of the embedding and add invariance properties to the detection, we deterministically compute a few augmentations on the cropped image. In this work, we use clockwise and anti-clockwise $90\degree$ rotations as well as horizontal flips of the cropped image.

Finally, the original image and the augmentations are fed into the image encoder part of CLIP to get the raw embeddings $v_{ij}$, where $i$ is the class and $j$ the $j$-th embedding of this class. The final embeddings passed into clip are computed as $z_i = normalize(\sum_j v_{ij})$, where $normalize$ normalizes the vector such that $\Vert z_i \Vert_2 = 1$. Several different images of a given target object can be processed in the same manner to get a single embedding vector for this object, all of the raw embeddings being aggregated together before the normalization step. The whole process of computing the final embedding vectors constitute the "training" of the model to recognize the new classes.


\subsection{Inference}

The final model can then be used in the same manner as the original YOLO-World model, but with the $z_i$ computed from the target object images instead of being computed from text prompts. Storing the raw vectors $v_{ij}$ is sufficient and still allows new images to be added for a given object if desired. The target object RGB images can be discarded as the embedding vectors already store all of the relevant information for the detection. The number of training images, including the images created by data augmentation add a computational cost at training time, but no additional storage and computational costs at inference time. The training and inference step are visually summarized in Figure \ref{Fig:ourmodel}.

It is worth noting that YOLO-World accepts any number of target embedding vectors $z_i$ as input at inference time, allowing the detection of different objects in a single forward pass and making the inference process very efficient. Due to the non-trivial interactions between embeddings in the attention layers of the RepVL-PAN module, the output of the YOLO-World model when specifying all of the embeddings vectors as input is generally not identical to the case where the embeddings vectors are independently presented one by one on the same query image.

\section{Experiments}
This section presents our experiments on the challenging Teachable Egocentric Objects Dataset (TEgO) \cite{lee2019hands}.
The code for all of our experiments is available at \url{https://anonymous.4open.science/r/yolo-tdid-C21B/README.md}.

\subsection{Object recognition on TEgO}
TEgO presents 19 objects, each photographed around 30 times in different settings.
The training images can appear in cluttered or uncluttered environments, with a hand in the frame or not, and the photograph taken by a person with a visual impairment or not. The testing images can be taken in cluttered or uncluttered environments, with or without camera flash and with or without indoor lighting.
The goal of this dataset is to allow the evaluation of TDID algorithms that have to classify novel objects using a small number of training examples and adds the difficult context of having half of the pictures taken by people with visual disabilities, which might thus make the objects appear blurry or partially occluded. This make this dataset a good proxy for estimating the performance of object detectors used by visually impaired people.

\subsection{Protocol}

For the experiments we follow the method explained in Section \ref{method}. 
We filter the images to keep only training images featuring the objects captured in an uncluttered environment by a sighted person, but keep all test images for evaluation. We do this to simulate having the best condition when saving the objects of interest since in practice we can detect if the object has been incorrectly captured (if the object appears blurry or if it is not well centered in the frame).
We evaluate the three sizes of the YOLO-World model $s$, $m$ and $l$ for small, medium and large respectively. We take the number of objects $c$ evaluated at the same time in the set $\{2, 4, 9, 19\}$ and the number of training examples per object $k$ in the set $\{1, 3, 5, 10\}$.
For each combination, we repeat the experiment $30$ times with a different random seed. We fix the cropping margin to $15$ pixels for the remaining of the paper.

For the evaluation, we only keep all test images, which include cluttered images and other difficult environmental effects. The train and test set images are distinct, but contain the same objects in different poses and environments.

We implement our experiments in PyTorch and run them on a Linux machine with a Nvidia RTX A$6000$ GPU and a Intel(R) Xeon(R) Gold $6240$R CPU. The evaluation of a model on TEgO takes about $1$ minutes, but this time varies depending on the number of training examples and classes considered. The evaluation of a batch of $3 \times 4 \times 4 \times 30 = 1440$ runs takes about $20$ hours.

\subsection{Results}

We report the accuracy metric by model size ($s$, $m$ and $l$ for small, medium and large respectively), number of classes, number of training examples and use of data augmentation in Table \ref{tab:tego_results}. Overall, the accuracy decreases with the number of total classes to detect, but increases with the model size and the number of examples. We also observe a slight but consistent increase ($\approx 3\%$ overall) in performance when using augmented examples images.
The standard deviations (indicated by parenthesis) are much higher when the number of classes is small, we find this phenomenon to be caused by the random sampling of pairs of classes that are very difficult to discriminate and greatly increase the error. This effect decreases as these difficult classes have a greater chance to be selected when the number of classes considered increases to the maximum of $19$.
Interestingly, the accuracy does not always increase monotonically with the number of example images. Beyond $3$ training examples, the accuracy barely increases and sometimes even decreases.

We also note that the accuracy globally increases by about one point if we filter out the images taken by a blind person from the test set, and by $3$ additional points if we also remove cluttered images, which remove parasitic objects that might be confused with the objects to recognize (we do not report the details of these results here). This indicates that even though special care has to be taken when capturing the target objects for training to avoid capturing the wrong object, visual impairments only marginally impact the final accuracy. This in turn suggests that visually impaired people can autonomously use this system nearly optimally.

\begin{table}
\centering
\setlength{\extrarowheight}{0pt}
\addtolength{\extrarowheight}{\aboverulesep}
\addtolength{\extrarowheight}{\belowrulesep}
\setlength{\aboverulesep}{0pt}
\setlength{\belowrulesep}{0pt}
\caption{Accuracy scores on TEgO}
\label{tab:tego_results}
\resizebox{1.0\columnwidth}{!}{%
\begin{tabular}{rrrrr|rrr} \toprule
\multicolumn{1}{l}{}                                                           & \multicolumn{1}{l}{}                                                            & \multicolumn{3}{c|}{Without augmentations}                                                                                                                      & \multicolumn{3}{c}{With augmentations}                                                                                                                                             \\
\multicolumn{1}{l}{}                                                           & \multicolumn{1}{l}{model size}                                                  & \multicolumn{1}{c}{s}                               & \multicolumn{1}{c}{m}                               & \multicolumn{1}{c|}{l}                              & \multicolumn{1}{c}{s}                               & \multicolumn{1}{c}{m}                                        & \multicolumn{1}{c}{l}                                         \\
\multicolumn{1}{l}{\begin{tabular}[c]{@{}l@{}}number of\\classes\end{tabular}} & \multicolumn{1}{l}{\begin{tabular}[c]{@{}l@{}}number of\\examples\end{tabular}} & \multicolumn{1}{c}{}                                & \multicolumn{1}{c}{}                                & \multicolumn{1}{c|}{}                               & \multicolumn{1}{c}{}                                & \multicolumn{1}{c}{}                                         & \multicolumn{1}{c}{}                                          \\ \midrule
\multirow{4}{*}{2}                                                             & 1                                                                               & 64.66\% (14.96)                                     & 72.41\% (15.67)                                     & 70.84\% (16.07)                                     & 67.36\% (15.13)                                     & 72.17\% (15.25)                                              & \textbf{75.65\%} (13.53)                                      \\
& {\cellcolor[rgb]{0.961,0.961,0.961}}3                                           & {\cellcolor[rgb]{0.961,0.961,0.961}}74.36\% (15.68) & {\cellcolor[rgb]{0.961,0.961,0.961}}73.06\% (14.81) & {\cellcolor[rgb]{0.961,0.961,0.961}}73.96\% (17.45) & {\cellcolor[rgb]{0.961,0.961,0.961}}75.92\% (15.05) & {\cellcolor[rgb]{0.961,0.961,0.961}}\textbf{78.93\%} (13.63) & {\cellcolor[rgb]{0.961,0.961,0.961}}76.90\% (15.93)           \\
& 5                                                                               & 69.82\% (15.18)                                     & 71.19\% (14.57)                                     & 75.39\% (11.18)                                     & 74.06\% (16.08)                                     & 73.79\% (14.74)                                              & \textbf{78.01\%} (17.05)                                      \\
& {\cellcolor[rgb]{0.961,0.961,0.961}}10                                          & {\cellcolor[rgb]{0.961,0.961,0.961}}71.54\% (15.01) & {\cellcolor[rgb]{0.961,0.961,0.961}}76.65\% (14.58) & {\cellcolor[rgb]{0.961,0.961,0.961}}79.23\% (12.92) & {\cellcolor[rgb]{0.961,0.961,0.961}}74.52\% (16.07) & {\cellcolor[rgb]{0.961,0.961,0.961}}71.92\% (15.64)          & {\cellcolor[rgb]{0.961,0.961,0.961}}\textbf{85.04\%} (10.87)  \\ \hline
\multirow{4}{*}{4}                                                             & 1                                                                               & 39.87\% (10.21)                                     & 47.75\% (12.30)                                     & 49.23\% (13.95)                                     & 46.29\% (11.16)                                     & 52.06\% (12.48)                                              & \textbf{54.20\%} (10.54)                                      \\
& {\cellcolor[rgb]{0.961,0.961,0.961}}3                                           & {\cellcolor[rgb]{0.961,0.961,0.961}}51.57\% (12.46) & {\cellcolor[rgb]{0.961,0.961,0.961}}51.56\% (13.23) & {\cellcolor[rgb]{0.961,0.961,0.961}}54.23\% (14.37) & {\cellcolor[rgb]{0.961,0.961,0.961}}50.04\% (11.33) & {\cellcolor[rgb]{0.961,0.961,0.961}}55.27\% (10.74)          & {\cellcolor[rgb]{0.961,0.961,0.961}}\textbf{59.93\%} (12.55)  \\
& 5                                                                               & 49.03\% (13.42)                                     & 53.77\% (13.31)                                     & 49.83\% (11.20)                                     & 49.38\% (12.64)                                     & 55.68\% (11.67)                                              & \textbf{59.20\%} (12.56)                                      \\
& {\cellcolor[rgb]{0.961,0.961,0.961}}10                                          & {\cellcolor[rgb]{0.961,0.961,0.961}}45.84\% (13.74) & {\cellcolor[rgb]{0.961,0.961,0.961}}52.51\% (11.30) & {\cellcolor[rgb]{0.961,0.961,0.961}}53.63\% (12.76) & {\cellcolor[rgb]{0.961,0.961,0.961}}49.49\% (12.02) & {\cellcolor[rgb]{0.961,0.961,0.961}}52.50\% (8.77)           & {\cellcolor[rgb]{0.961,0.961,0.961}}\textbf{64.38\%} (11.91)  \\ \hline
\multirow{4}{*}{9}                                                             & 1                                                                               & 24.45\% (6.24)                                      & 27.43\% (6.39)                                      & 27.93\% (6.78)                                      & 24.88\% (6.37)                                      & 30.89\% (5.85)                                               & \textbf{35.91\%} (7.87)                                       \\
& {\cellcolor[rgb]{0.961,0.961,0.961}}3                                           & {\cellcolor[rgb]{0.961,0.961,0.961}}25.20\% (7.65)  & {\cellcolor[rgb]{0.961,0.961,0.961}}30.38\% (6.07)  & {\cellcolor[rgb]{0.961,0.961,0.961}}31.72\% (6.49)  & {\cellcolor[rgb]{0.961,0.961,0.961}}27.35\% (6.01)  & {\cellcolor[rgb]{0.961,0.961,0.961}}33.96\% (6.03)           & {\cellcolor[rgb]{0.961,0.961,0.961}}\textbf{36.78\%} (6.52)   \\
& 5                                                                               & 25.78\% (6.38)                                      & 30.74\% (5.88)                                      & 31.63\% (4.95)                                      & 29.36\% (5.68)                                      & 35.29\% (6.70)                                               & \textbf{38.02\%} (6.30)                                       \\
& {\cellcolor[rgb]{0.961,0.961,0.961}}10                                          & {\cellcolor[rgb]{0.961,0.961,0.961}}27.03\% (8.13)  & {\cellcolor[rgb]{0.961,0.961,0.961}}30.32\% (5.28)  & {\cellcolor[rgb]{0.961,0.961,0.961}}35.28\% (7.20)  & {\cellcolor[rgb]{0.961,0.961,0.961}}29.53\% (5.63)  & {\cellcolor[rgb]{0.961,0.961,0.961}}36.11\% (6.50)           & {\cellcolor[rgb]{0.961,0.961,0.961}}\textbf{39.64\%} (5.74)   \\ \hline
\multirow{4}{*}{19}                                                            & 1                                                                               & 11.62\% (2.51)                                      & 15.53\% (3.09)                                      & 17.71\% (2.58)                                      & 14.12\% (2.69)                                      & 16.82\% (3.01)                                               & \textbf{20.06\%} (3.42)                                       \\
& {\cellcolor[rgb]{0.961,0.961,0.961}}3                                           & {\cellcolor[rgb]{0.961,0.961,0.961}}14.36\% (2.18)  & {\cellcolor[rgb]{0.961,0.961,0.961}}16.51\% (1.99)  & {\cellcolor[rgb]{0.961,0.961,0.961}}18.25\% (2.49)  & {\cellcolor[rgb]{0.961,0.961,0.961}}15.63\% (1.97)  & {\cellcolor[rgb]{0.961,0.961,0.961}}20.60\% (2.41)           & {\cellcolor[rgb]{0.961,0.961,0.961}}\textbf{22.81\%} (2.44)   \\
& 5                                                                               & 13.87\% (2.04)                                      & 17.28\% (1.44)                                      & 17.94\% (1.66)                                      & 16.75\% (1.69)                                      & 20.14\% (1.57)                                               & \textbf{23.48\%} (2.05)                                       \\
& {\cellcolor[rgb]{0.961,0.961,0.961}}10                                          & {\cellcolor[rgb]{0.961,0.961,0.961}}13.38\% (1.28)  & {\cellcolor[rgb]{0.961,0.961,0.961}}17.45\% (1.18)  & {\cellcolor[rgb]{0.961,0.961,0.961}}18.52\% (1.20)  & {\cellcolor[rgb]{0.961,0.961,0.961}}16.71\% (1.37)  & {\cellcolor[rgb]{0.961,0.961,0.961}}21.66\% (1.56)           & {\cellcolor[rgb]{0.961,0.961,0.961}}\textbf{23.73\%} (1.15)   \\ \bottomrule
\end{tabular}%
}
\end{table}

\begin{figure}[t]
\centering
\includegraphics[width=1.0\textwidth]{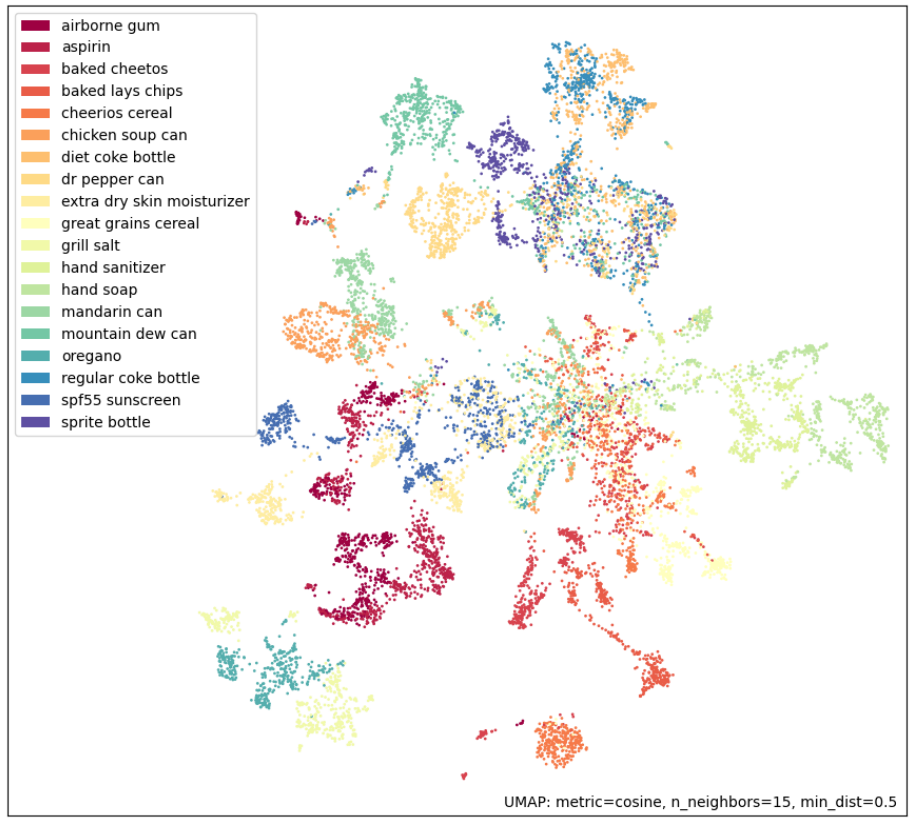}
\caption{UMAP plot of the CLIP embeddings computed from the cropped TEgO images.}\label{Fig:umap}
\end{figure}

\begin{figure}[t]
\centering
\includegraphics[width=1.0\textwidth]{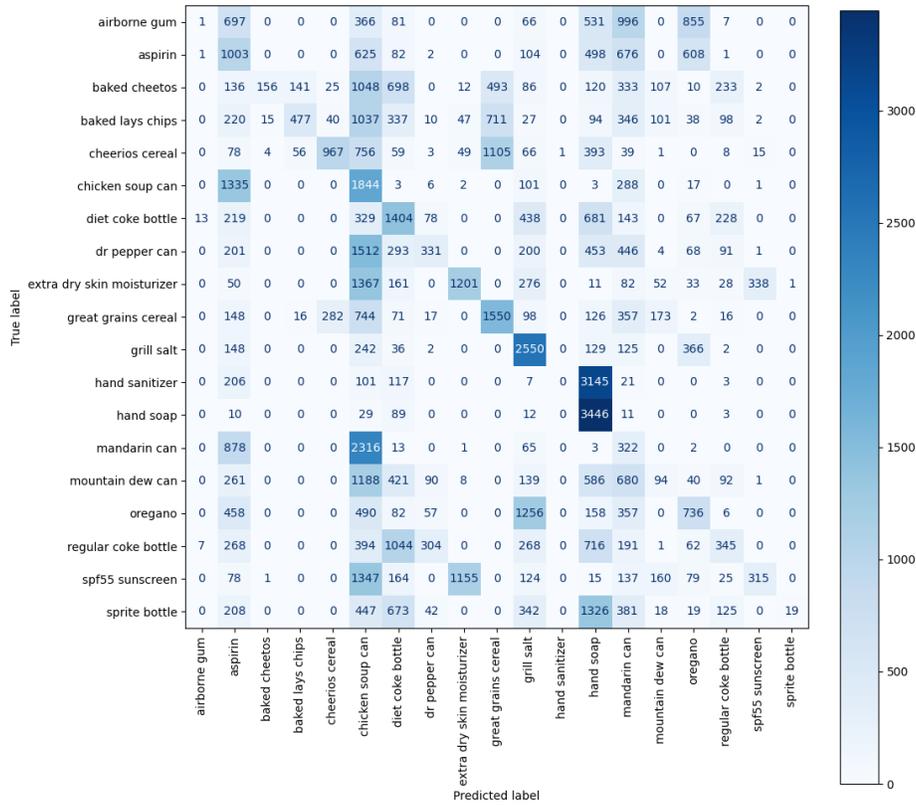}
\caption{Confusion matrix on the test set}\label{Fig:confusion}
\end{figure}

\subsection{Analysis}

In order to better understand where and why the models make classification errors, we compute a 2D UMAP \cite{sainburg2021parametric} projection of the cropped image embeddings using the cosine metric, as well as the confusion matrix on the test set. The projection shown in Figure \ref{Fig:umap} shows that a few classes are well separated from the rest, but there is also a large amount of overlap in some regions of the space, indicating that the embeddings computed by CLIP are not themselves sufficiently clustered, as confirmed by a silhouette score of about $0.06$. If we  keep only the cropped images from uncluttered environment taken by sighted people that contain very salient objects, the silhouette score increases to around $0.33$, indicating that a significant amount of overlap is likely to be caused by the first step of our method, that is localizing and cropping the object of interest from the  training image. Indeed, if we randomly visually examine image crops, we quickly find bad crops, where either the wrong object from the clutter or only a part of the object of interest has been captured.

We then compute the confusion matrix in Figure \ref{Fig:confusion}. The confusion matrix reveals two important information. Firstly, some objects are sometimes confused with other objects, as illustrated by the ``hand sanitizer'' object that is almost always predicted as the ``hand soap'', these objects being extremely visually similar.
Secondly, a few objects such as the ``chicken soup can'' appear to absorb the predictions of other classes. Indeed, even visually very distinct objects are predicted as this class more often than the correct class. Looking back at Figure \ref{Fig:umap}, we find that ``chicken soup can'' embeddings are indeed close to ``mandarin can'' embeddings, but do not particularly overlap with the other classes. This explains why the model classifies one of these objects into the other class, but does not explain why unrelated objects are predicted as the ``chicken soup can'' class. This loss in performance might be the result of a problem with the image embeddings, or a limitation of the trained RepVL-PAN module itself.

\subsection{Improving the Results using Whitening and Coloring}

We conjecture that a distributional shift occurs between the text and image embeddings. Indeed, YOLO-World is pre-trained with text embeddings that do not necessarily share the same statistics as image embeddings, even though they share a common latent space in which matching image-text pairs have high dot products. If these distribution statistics do not match, this is essentially the same as using YOLO-World with Out-Of-Distribution (OOD) data. If the model is not robust OOD, a loss of performance is to be expected.

In order to test this conjecture and improve accuracy, we propose a method to reduce the distributional shift by transforming the image embeddings so that their mean and covariance statistics match that of the text embeddings and thus provide the RepVL-PAN module with inputs closer to the original training distribution.

\subsubsection{Whitening and Coloring.}

The transformation we chose is a \textit{Whitening} transformation, more specifically a Zero-phase Component Analysis (ZCA) or Mahalanobis transformation, followed by a \textit{Coloring} transformation using the target covariance matrix. More specifically, if $\Sigma_\text{img}$ if the covariance matrix of image embeddings, the whitening matrix is $W_{\text{ZCA}}=E_\text{img} D_\text{img}^{-\frac{1}{2}} E_\text{img}^T$ where $E_\text{img}$ are the eigenvectors and $D_\text{img}$ the eigenvalues of $\Sigma_\text{img}$. To apply the whitening on the centered embedding vectors $X$, we use a matrix product $X_W= X W_\text{ZCA}$.
The coloring step is a similar procedure applied in reverse, we compute the coloring matrix as $W_\text{color}=E_\text{txt} D_\text{txt}^{\frac{1}{2}} E_\text{txt}^T$ with respect to the target covariance matrix $\Sigma_\text{txt}$ and obtain the final transformed embeddings as $X_\text{WC} = X_W W_\text{color} + \mu_\text{txt}$, where $\mu_\text{txt}$ is the estimated mean of the text embeddings. After applying the full transformation, the resulting embeddings $X_\text{WC}$ have approximately the mean and covariance statistics of the text embedding distribution. This transformation was chosen because of its simplicity and ability to preserve the original basis, which is necessary for the RepVL-PAN module to function correctly as it expects embedding vectors in a specific basis. The full transformation adds a small computational cost that is equivalent to adding a single dense layer to the end of the CLIP image encoder when computing image embeddings for the few-shot training step. There is no additional cost at inference time.

We compute text statistics using singularized noun classes from the GQA dataset~\cite{hudson2019gqa}, and image statistics using training images from COCO~$2017$~\cite{lin2014microsoft}. We use the CLIP text and image encoders to compute the text and image embeddings respectively. Embeddings are normalized by their $L^2$ norm before the computations of the statistics since text embeddings vectors were normalized for the original training of YOLO-World. These datasets were chosen because their distribution are similar to the image and text distributions used to train the YOLO-World models.

\subsubsection{Results.}

Table \ref{tab:wc} reports the accuracy we obtain as a result of using both data augmentation and Whitening-Coloring method. Standard deviations are specified between parenthesis and the right side of the table shows improvement over the results reported in Table \ref{tab:tego_results} with data augmentation only.
The use of this additional transformation mostly improves the accuracy for the small and medium YOLO-World models, as well as for large number of classes. We notice a few extreme changes, for instance accuracy increases by more than $8$ points for the medium model with $2$ classes and $10$ training examples, but decreases by almost $8$ points for the large model. In Table \ref{tab:tego_results}, these values corresponded to abnormally low and and high accuracy respectively compared to the results obtained with the same number of classes but different number of training examples. Finally, when the classification task includes all of the $19$ classes, the highest accuracy is now $26.23\%$ for the large model with $10$ training examples, while the highest accuracy was previously $23.73\%$. This increase of $2.5$ points in accuracy is essentially free given the negligible computational cost of the whitening and coloring step.

\begin{table}
\caption{Results after adding image embedding transformation using whitening and coloring matrices.}\label{tab:wc}
\centering
\resizebox{1.0\columnwidth}{!}{%
\begin{tabular}{lrrrr|>{\centering\arraybackslash}p{2cm}>{\centering\arraybackslash}p{2cm}>{\centering\arraybackslash}p{2cm}} \hline
   & \multicolumn{1}{l}{}                                                            & \multicolumn{3}{c|}{Whitening-Coloring
and augmentations}            & \multicolumn{3}{c}{Improvement
over only augmentations}              \\
   & \multicolumn{1}{l}{model size}                                                  & \multicolumn{1}{c}{s} & \multicolumn{1}{c}{m} & \multicolumn{1}{c|}{l} & \multicolumn{1}{c}{s} & \multicolumn{1}{c}{m} & \multicolumn{1}{c}{l}  \\
\begin{tabular}[c]{@{}l@{}}number of\\classes\end{tabular} & \multicolumn{1}{l}{\begin{tabular}[c]{@{}l@{}}number of\\examples\end{tabular}} & \multicolumn{1}{c}{}  & \multicolumn{1}{c}{}  & \multicolumn{1}{c|}{}  & \multicolumn{1}{c}{}  & \multicolumn{1}{c}{}  & \multicolumn{1}{c}{}   \\ \hline
\multicolumn{1}{r}{2}                                      & 1                                                                               & 68.68\% (16.08)       & 73.55\% (17.30)       & 81.77\% (15.64)        & +1.31                 & +1.38                 & +6.12                  \\
\rowcolor[rgb]{0.961,0.961,0.961}                          & 3                                                                               & 74.26\% (17.36)       & 72.02\% (16.28)       & 75.42\% (14.04)        & -1.66                 & -6.91                 & -1.48                  \\
& 5                                                                               & 74.16\% (17.91)       & 76.97\% (14.14)       & 80.08\% (13.13)        & +0.10                 & +3.19                 & +2.07                  \\
\rowcolor[rgb]{0.961,0.961,0.961}                          & 10                                                                              & 75.19\% (14.25)       & 80.38\% (13.84)       & 77.32\% (15.69)        & +0.67                 & +8.46                 & -7.71                  \\ \hline
\multicolumn{1}{r}{4}                                      & 1                                                                               & 48.89\% (12.53)       & 49.12\% (11.20)       & 53.75\% (13.68)        & +2.60                 & -2.93                 & -0.45                  \\
\rowcolor[rgb]{0.961,0.961,0.961}                          & 3                                                                               & 49.48\% (11.19)       & 54.92\% (7.91)        & 58.82\% (12.52)        & -0.56                 & -0.35                 & -1.11                  \\
& 5                                                                               & 52.79\% (11.64)       & 56.35\% (14.61)       & 57.00\% (11.92)        & +3.41                 & +0.68                 & -2.20                  \\
\rowcolor[rgb]{0.961,0.961,0.961}                          & 10                                                                              & 51.09\% (11.78)       & 56.22\% (9.45)        & 57.50\% (11.03)        & +1.60                 & +3.72                 & -6.87                  \\ \hline
\multicolumn{1}{r}{9}                                      & 1                                                                               & 27.98\% (7.73)        & 31.62\% (6.27)        & 34.60\% (7.53)         & +3.10                 & +0.73                 & -1.30                  \\
\rowcolor[rgb]{0.961,0.961,0.961}                          & 3                                                                               & 32.78\% (5.63)        & 36.02\% (6.11)        & 36.62\% (5.91)         & +5.43                 & +2.06                 & -0.17                  \\
& 5                                                                               & 31.68\% (4.95)        & 37.02\% (7.18)        & 37.34\% (6.02)         & +2.32                 & +1.73                 & -0.67                  \\
\rowcolor[rgb]{0.961,0.961,0.961}                          & 10                                                                              & 32.33\% (5.69)        & 38.89\% (6.98)        & 39.63\% (5.09)         & +2.80                 & +2.78                 & -0.00                  \\ \hline
\multicolumn{1}{r}{19}                                     & 1                                                                               & 15.43\% (3.13)        & 18.14\% (2.57)        & 21.31\% (2.98)         & +1.31                 & +1.32                 & +1.25                  \\
\rowcolor[rgb]{0.961,0.961,0.961}                          & 3                                                                               & 18.64\% (2.62)        & 21.32\% (2.42)        & 23.89\% (2.08)         & +3.01                 & +0.72                 & +1.08                  \\
& 5                                                                               & 18.04\% (1.32)        & 21.67\% (1.61)        & 25.53\% (2.28)         & +1.28                 & +1.53                 & +2.05                  \\
\rowcolor[rgb]{0.961,0.961,0.961}                          & 10                                                                              & 17.94\% (1.26)        & 22.23\% (1.44)        & 26.23\% (1.82)         & +1.23                 & +0.58                 & +2.50                  \\ \hline
\end{tabular}}
\end{table}

\begin{figure}[tb]
  \centering
  \begin{subfigure}{0.45\linewidth}
    \includegraphics[width=1.0\textwidth]{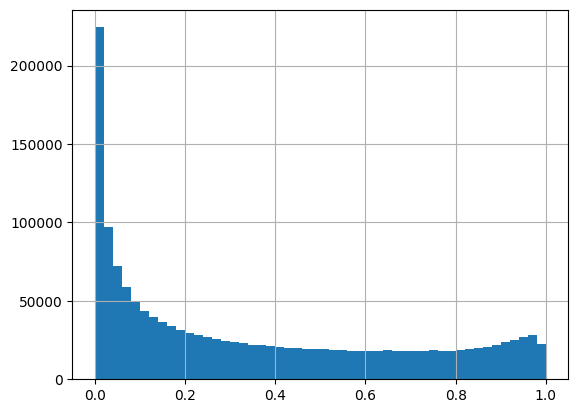}
    \caption{without whitening and coloring}
    \label{fig:hist-a}
  \end{subfigure}
  \hfill
  \begin{subfigure}{0.45\linewidth}
    \includegraphics[width=1.0\textwidth]{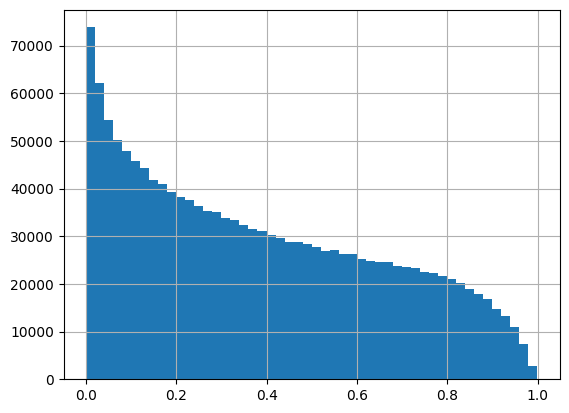}
    \caption{with whitening and coloring}
    \label{fig:hist-b}
  \end{subfigure}
  \caption{Histograms of the predicted probabilities for incorrect classifications}
  \label{fig:hist}
\end{figure}

We also compute the histograms of predicted probabilities of incorrect classifications with and without whitening and coloring in Figure \ref{fig:hist}. As reported in Figure \ref{fig:hist-a}, without the statistical adjustment the predicted probabilities for incorrect classes are often very close to $0$, but a non-negligible of predictions saturate to a probability of $1$, which demonstrates the over-confidence of the model. At the contrary, as Figure \ref{fig:hist-b} shows, the predicted probabilities follow a more expected monotonically decreasing density curve when statistical adjustments are made to the embeddings to make them follow a distribution closer to the one used while pre-training the model. In this case only small fractions of incorrectly classified examples have been predicted with probabilities close to $1$ by the models.

Our whitening and coloring transformation thus not only improves the accuracy, it also helps the model to predicted calibrated class probabilities. This fact is particularly important for having the ability to reject insufficiently confident classifications using a threshold with deployed models.




\section{Discussion}


We have described a method to convert reparametrizable open-vocabulary object detection models into TDID models and applied it to the YOLO-World model. We then evaluated this model as a few-shot object detector on the challenging TEgO dataset.

\subsection{Comparison with State of the Art}

As very little work has previously been done on TDID, we cannot present meaningful comparison experiments with other state of the art techniques.
The T-Rex2 model \cite{jiang2024trex2} is the only published method which supports TDID, but the model is behind an API, so we consider this model ineligible for scientific experiments. The T-Rex2 model also needs a unique input bounding box to specify the object to detect, and can only detect this object in a single forward pass. At the contrary, our method automatically detects and crops the object of interest, allows multiple image examples as input, and can detect any number of saved object instances in a single forward pass.

Compared to \cite{Mercier2021}, our work is concerned with only real images, some of which contain clutter. Our method also does not require to separate the object instances from the background and does not require any additional training of the neural network weights. We re-use a model with a simpler architecture, which can still double as an open-set object detection model using user specified classes in the form of natural text, making our method more general and flexible.
Due to the absence of code, unavailability of the model weights and the difficulty of adapting their synthetic pipeline to the TEgO dataset, we chose leave this work out of our experiments as well.

\subsection{Deployment of the Solution}

Overall, we believe our work is a step towards the creation of smart assistant with all computations being done on-device. Our method is general and compatible with some existing pre-trained models and shows encouraging performance on real-world images. Thanks to our few-shots learning algorithm, we do not require long and expensive on-device fine-tuning of the model parameters at all and makes the use of the method compatible with existing deep learning mobile inference engines such as Tensorflow~\cite{tensorflow2015-whitepaper} Lite, PyTorch Mobile\footnote{\url{https://pytorch.org/mobile/home/}} or ExecuTorch\footnote{\url{https://pytorch.org/executorch-overview}}.

\subsection{Object Detection Performance}

Even though our method is advantageous in various regards, it is still far from perfect. Firstly, all of the YOLO-World models react strangely for some saved object examples for no clear reason why, which greatly decreases the accuracy. In principle this problem might be mitigated by ensuring saved objects are not confused with already saved objects in the training phase and perform the detection step separately for the problematic objects if needed.
Secondly, the performance of the object detection largely hinges on the ability of the model to accurately detect and crop the training objects in the first place, which correspond to the first step of the training phase. We manually inspect the image crops and identify instances where only a small part of the object is captured, for instance when object images are themselves printed on the object of interest (e.g. a cereal box with a bowl printed on it). There are also some instances where the object was completely missed and the resulting cropped image was a part of the background.

We conjecture that the problem of having objects that confuse the model comes from a distributional shift between text and image embeddings, to which the YOLO-World model is not robust. Indeed, even though the text and image embeddings were obtained by maximizing the dot product for matching pairs, the text labels used for training YOLO-World follow a different distribution from the data distribution of captions in the dataset used to train CLIP.
To alleviate this problem, it might be necessary to fine-tune the Rep-VL-PAN module or the whole model on matching image pairs. However, we risk losing the open-vocabulary detection abilities of the model because of the new distributional shift caused by this fine tuning.

\section{Conclusion}

This paper showed how to transform YOLO-World, an open-vocabulary detection model into a target driven detection model. This transformation is enabled by the fact that the original model uses text embedding vectors produced by CLIP, a contrastive language-image model which can also be used to encode images into vectors, replacing the original vocabulary. The re-purposed models achieve reasonable one-shot and few-shot classification performance on TEgO, a challenging object recognition dataset partially produced by visually impaired people. Using more than one image as example increases the classification accuracy, but the gain is marginal beyond $3$ image examples per object. The use of deterministic image augmentation slightly improves the accuracy for all model sizes and number of objects to detect.

We conjecture that image embedding statistics may differ from text embedding statistics in a way that might harm the classification performance. To reduce the importance of this distributional shift, we use a whitening transformation followed by a coloring transformation to adjust the mean and covariance statistics of the image embeddings to that of normalized text embeddings. We show that this simple and cost-efficient transformation indeed increases performance further, especially when the number of classes is large. The adjustment of image embeddings statistics also has the benefit of offering better calibrated predicted class probabilities.

Our method enables the deployment of edge-device object detectors that can be continually re-trained on-device with personal objects at a small computational and storage cost.
In our future work, we will try to obtain a higher TDID performance by fine-tuning the resulting model or just the RepVL-PAN module to increase the few-shot recognition performance of the model when using image embeddings. 

%
%
\bibliographystyle{splncs04}
\bibliography{main}
\end{document}